\documentclass{article} % For LaTeX2e
\usepackage{nips14submit_e,times}
\usepackage{hyperref}
\usepackage{url}
\usepackage{algpseudocode}
\usepackage{algorithm}
\usepackage{amssymb}
\usepackage{amsthm}
\usepackage{amsmath}
\usepackage{graphicx}
\usepackage{colonequals}
\usepackage[section]{placeins}
\DeclareGraphicsExtensions{.pdf,.png,.jpg}
\usepackage{booktabs}
\usepackage{array}

\title{Bandit Label Inference for \\Weakly Supervised Learning}

\author{
Ke Li \qquad Jitendra Malik \\
Department of Electrical Engineering and Computer Sciences\\
University of California, Berkeley\\
Berkeley, CA 94720\\
United States\\
\texttt{\{ke.li,malik\}@eecs.berkeley.edu} \\
}

\nipsfinalcopy % Uncomment for camera-ready version

\begin{document}

\maketitle

\begin{abstract}
The scarcity of data annotated at the desired level of granularity is a recurring issue in many applications. Significant amounts of effort have been devoted to developing weakly supervised methods tailored to each individual setting, which are often carefully designed to take advantage of the particular properties of weak supervision regimes, form of available data and prior knowledge of the task at hand. Unfortunately, it is difficult to adapt these methods to new tasks and/or forms of data, which often require different weak supervision regimes or models. We present a general-purpose method that can solve any weakly supervised learning problem irrespective of the weak supervision regime or the model. The proposed method turns any off-the-shelf strongly supervised classifier into a weakly supervised classifier and allows the user to specify any arbitrary weakly supervision regime via a loss function. We apply the method to several different weak supervision regimes and demonstrate competitive results compared to methods specifically engineered for those settings. 
\end{abstract}

\section{Introduction}

The problem of weakly supervised learning naturally arises in a wide range of domains, including computer vision and natural language processing. Current weakly supervised methods are typically engineered specifically for particular weak supervision regimes, such as multiple instance learning (MIL) \cite{dietterich1997solving}, learning with label proportions (LLP) \cite{defreitas2005learning} or other domain-specific settings \cite{huang2014learning}, and are often carefully designed to take advantage of the form of available data and prior knowledge of the task at hand to maximize the amount of supervisory signal. These approaches usually take the form of extensions of models that are known to work well in the strongly supervised setting and necessitate the design of clever heuristics to ensure the quality of the resulting locally optimal solutions. While they represent sensible approaches, it is difficult to adapt them to new tasks and/or forms of data, which often require different weak supervision regimes or models. As a result, performance gains in one weakly supervised setting do not generally translate to performance gains in other settings. 

In this paper, we propose a general-purpose method for weakly supervised learning that is able to solve any weakly supervised learning problem. Unlike existing methods, the proposed method is agnostic to the regime of weak supervision and the choice of the model. It reduces any weakly supervised learning problem to a strongly supervised learning problem and enables the use of any strongly supervised learning algorithm in the weakly supervised setting. It supports any weakly supervised regime by representing it as a user-defined loss function, and thus can serve as a drop-in replacement for any existing weakly supervised algorithm. The proposed method works by first inferring the instance-level labels and then training a classifier on the inferred labels in a fully supervised manner. Labels are inferred in an efficient manner using a combinatorial multi-armed bandit algorithm; for this reason we dub the proposed method Bandit Label Inference as Supervisory Signal, or BLISS for short. We apply the method to various weak supervision regimes and show competitive empirical results compared to methods specifically designed for those settings. 

\section{Related Work}

There has been a rich literature of work investigating different ways of relaxing the level of supervision required to learn a model. Perhaps the most extensively studied setting is the multiple instance learning (MIL) regime, where the objective is to train a classifier from binary labelled bags of unlabelled instances, with positive bags known to contain at least one positive instance and negative bags containing no positive instances. Various algorithms have been developed for the MIL regime, including Axis-Parallel Rectangle (APR) \cite{dietterich1997solving}, Diverse Density \cite{maron1998framework} and EM-DD \cite{zhang2001dd}. Many other MIL algorithms take the form of extensions of strongly supervised methods, such as $k$-nearest neighbour methods Citation-kNN and Bayesian-kNN \cite{wang2000solving}, support vector machine methods mi-SVM and MI-SVM \cite{andrews2002} and neural network algorithm BP-MIP \cite{zhang2004improve}. The MIL setting has been found to arise naturally in a wide range tasks in various domains, such as drug activity prediction \cite{dietterich1997solving}, stock selection \cite{maron1998framework}, text categorization \cite{andrews2002}, object detection \cite{song2014learning} and computer-aided medical diagnosis \cite{fung2007multiple}. Various extensions of the MIL regime have also been explored, such as settings where the bag label depends on all instances in the bag \cite{xu2004logistic} or where the bag label is positive only when multiple conditions are simultaneously satisfied \cite{weidmann2003two}. 

One other notable setting that has been studied is the learning with label proportions (LLP) regime, where the bag label is the proportion of positive instances in the bag. A variety of methods have been developed, such as approaches based on graphical model formulations \cite{defreitas2005learning}, $k$-means based methods \cite{chen2009kernel}, support vector machines \cite{rueping2010svm,yu2013svm}, and estimations of the mean operator \cite{quadrianto2008estimating}. The LLP regime has found applications in fraud detection \cite{rueping2010svm} and video event detection \cite{lai2014video}. 

\section{Bandit Label Inference as Supervisory Signal}

A weakly supervised learning problem arises when some or all labels of individual training instances are unknown. So, if the labels of the training instances can be inferred, this problem is reduced to the standard strongly supervised learning setting. All weakly supervised regimes fit nicely into this framework, and different ways of leveraging weak labels can be represented as different loss functions on the inferred instance-level labels given the weak labels. 

The problem of inferring instance-level labels is challenging, as the number of possible labellings scales exponentially in the number of instances. The labels of different instances are highly dependent, so it is not possible to optimize over the labels of each instance independently. We tackle this problem by formulating the label inference problem as a combinatorial multi-armed bandit (CMAB) problem and leveraging a CMAB algorithm to explore the labelling space efficiently. Our formulation treats the strongly supervised classifier and the loss function induced by the weak supervision regime as a black box, enabling the proposed algorithm to work with any combination of classifier and weak supervision regime. 

\subsection{Background}

The multi-armed bandit (MAB) is a general framework that models sequential decision-making under uncertainty. In its most basic form, there is a finite number of arms, each of which generates a reward from an unknown probability distribution when pulled. Only one arm can be pulled at a time and the objective is to choose the arm to pull in each round in order to maximize the cumulative expected reward one receives over multiple rounds. This is often equivalently formulated as minimization of cumulative pseudo-regret, which is defined as the difference in cumulative expected reward between the optimal and the actual arm selection strategy. Many arm selection strategies have been proposed; one classic strategy is the Upper Confidence Bound (UCB) strategy \cite{auer2002finite}, which computes a probabilistic upper bound for the true mean reward of each arm based on the sample mean and picks the arm with the highest upper bound. The probability at which the upper bound holds decreases over time, leading to a gradual transition from exploration (trying arms that have not been pulled much) to exploitation (pulling the arms that are known to give high rewards). When the reward distribution associated with each arm is in the exponential family, it has been shown that a generalization of the original UCB strategy, KL-UCB \cite{cappe2013kullback}, is optimal, in the sense that the leading term of the upper bound on cumulative pseudo-regret matches the lower bound \cite{lai1985asymptotically}. By considering the special case of Bernoulli-distributed rewards, an upper bound on cumulative pseudo-regret can be obtained for any reward distribution with bounded support on $[0,1]$, but it may not be optimal unless the rewards are Bernoulli. We refer interested readers to \cite{bubeck2012regret} for a survey on the topic. 

The combinatorial multi-armed bandit (CMAB) extends the classical MAB setting by allowing a set of (simple) arms, known as a super arm, to be pulled simultaneously. The super arms may have some underlying combinatorial structure; so, only some combinations of simple arms may be considered as valid super arms. The rewards of different simple arms in a super arm may be dependent, and the reward of a super arm can be thought of as the sum of the rewards of its constituent simple arms (though this can be generalized). Now, the objective is to choose a super arm to pull in each round to maximize the cumulative expected reward of super arms. Chen et al. \cite{chen2013} proposed an arm selection strategy for this setting called the Combinatorial Upper Confidence Bound (CUCB), which maintains a probabilistic upper bound for the true mean reward of each simple arm and picks the super arm with the highest sum of upper bounds of constituent simple arms. 

\subsection{Formulation}

In our formulation, we associate each instance with a set of simple arms, each of which corresponds to a possible label the instance can take. Assuming each instance only has one ground truth label, a super arm is valid if it consists of exactly one simple arm from each set associated with an instance. So, each super arm corresponds to a possible labelling of the instances. The reward of a super arm can be viewed as the negative loss on the labelling associated with the super arm. The form of the loss function depends on the weak supervision regime and task-specific prior knowledge, which impose hard and/or soft constraints on the space of the possible labellings; if a constraint is violated, loss should be high, or equivalently, reward should be low. This only needs to hold in expectation, since rewards can be stochastic. For example, in the MIL regime, we use a reward that penalizes labellings where no positive instances appear in a positive bag and a positive instance appears in a negative bag. The rewards of the simple arms are derived from the reward of the super arm, and are typically the local versions of the global reward of the super arm that serve as proxies for the marginal effect of assigning a particular label to an instance. 

By plugging in different reward functions, we can obtain algorithms that solve the label inference problem under different weak supervision regimes. Note that there are very few restrictions on the form of the reward function; in fact, almost any bounded loss function\footnote{Refer to \cite{chen2013} for the precise conditions on the reward function required by the CUCB algorithm.} can be turned into a valid reward function. 

\subsection{Algorithm}

We present the proposed label inference algorithm in detail below, which is based on the CUCB algorithm applied to our formulation. Conceptually, in each iteration of the algorithm, we first generate a candidate labelling of the training instances, which corresponds to pulling a super arm in the CMAB problem. Then, given the labelling, we train a classifier in a fully supervised manner and run the classifier on a weakly labelled held-out set. We obtain the predicted labels for each instance in the held-out set and compute the rewards, which is a function of the number of violations of the weak supervision constraints on the held-out set. Using this information, we can update the likelihood of each label, which corresponds to updating the empirical mean rewards of simple arms in the CMAB problem; the likelihood of the labels is then used to generate the candidate labelling in the subsequent iteration. The algorithm terminates after a fixed number of iterations and outputs the labels with the highest likelihood/empirical means. Please refer to Algorithm \ref{alg:bandit_label_inference} for a precise statement of the algorithm. 

\begin{algorithm}
\caption{Bandit label inference algorithm}
\label{alg:bandit_label_inference}
\begin{algorithmic}
\Require Set of possible labels each instance can take, reward $R_{x}(l)$ for assigning label $l$ to instance $x$ and strongly supervised classifier $f$
\Function{InferLabels}{$S$}
\State $T_{x,l} \gets 0$ and $s_{x,l} \gets 0$ for all instances $x \in S$ and all possible labels $l$ that $x$ can take
\State \Comment{$T_{x,l}$ is the number of times the simple arm $(x,l)$ is pulled and $s_{x,l}$ is the cumulative reward over all pulls}
\While{$\exists x_{0},l_{0}\; T_{x_{0},l_{0}} = 0$}
    \State Randomly pick a label assignment $\pi$ to all instances such that $\pi(x_{0}) = l_{0}$
    \State Train classifier $f$ on $S$ and $\pi(S)$ and get reward $R_{x}(\pi(x))$ for each instance $x$
    \State $\forall x \in S\; T_{x,\pi(x)} \gets T_{x,\pi(x)} + 1$
    \State $\forall x \in S\; s_{x,\pi(x)} \gets s_{x,\pi(x)} + R_{x}(\pi(x))$
\EndWhile
\For{$t \in [N]$}	\Comment{$\frac{s_{x,l}}{T_{x,l}}$ is the empirical mean reward}
    \State $\bar{\mu}_{x,l} \gets \frac{s_{x,l}}{T_{x,l}} + \sqrt{\frac{3\log t}{2T_{x,l}}}$ for all $x,l$	\Comment{Upper confidence bound for each simple arm}
    \State Pick the label assignment $\pi$ such that $\pi(x) = \arg\max_{l}\bar{\mu}_{x,l}$ for all $x$
    \State Train classifier $f$ on $S$ and $\pi(S)$ and get reward $R_{x}(\pi(x))$ for each $x$
    \State $\forall x \in S\; s_{x,\pi(x)} \gets s_{x,\pi(x)} + R_{x}(\pi(x))$ and $T_{x,\pi(x)} \gets T_{x,\pi(x)} + 1$
\EndFor
\State \Return label assignment $\pi^{*}$ such that $\pi^{*}(x) = \arg\max_{l}\frac{s_{x,l}}{T_{x,l}}$ for all $x$ and confidence scores $c$ such that $c(x) = \frac{s_{x,\pi^{*}(x)}}{T_{x,\pi^{*}(x)}} - \max_{l \neq \pi^{*}(x)}\frac{s_{x,l}}{T_{x,l}}$
\EndFunction
\end{algorithmic}
\end{algorithm}

Because the algorithm maintains a likelihood of each label for each instance, we can quantify the uncertainty of each inferred label by computing the difference in likelihood between the predicted label and its nearest competitor, which will be referred to as the confidence score. Using these confidence scores, we can devise a bootstrapping procedure, where the label inference algorithm is repeated multiple times with the most confident labels obtained in earlier passes serving as additional training data for the classifier in later passes. Confidence scores are also useful downstream when the inferred labels are further processed; for example, they can be used as weights when training the final classifier on the inferred labels. 

We also extend the algorithm to enable us to try multiple possible labellings in parallel. So, instead of picking a single super arm in each iteration, we pick a sequence of super arms to pull in parallel. Because we don't know the rewards we will receive from pulling the other super arms at the time of arm selection, each super arm is obtained by assuming the rewards that will be received from pulling super arms earlier in the sequence are equal to the empirical mean rewards observed so far. Empirically, we found that the set of super arms picked this way tends to be fairly diverse. 

Finally, in order to obtain inferred labels for all weakly-labelled instances in a dataset, we use a variant of cross-validation, i.e.: we divide the dataset into $K$ folds and use one fold as the training set and the other folds as the held-out set for the label inference algorithm. 

\subsection{Comparison with Existing Approaches}

The proposed method decouples the form of labelling constraints imposed by the weak supervision regime from the inner workings of the model -- any strongly supervised classifier can be fed in as a black box and the classifier does not need to know about the form of weak supervision. This flexibility offers several advantages: first, any strongly supervised classifier can be ported to the weakly supervised setting without modification, thereby enabling one to easily take advantage of advances made in the strongly supervised setting. Second, an algorithm that works well in a particular weakly supervised setting can be easily extended to work in a different weakly supervised setting by way of changing the rewards. Consequently, one can incorporate additional side information and/or prior knowledge on the labellings without needing to concern oneself with possible optimization challenges this would introduce. In addition, a weakly supervised algorithm that works in the binary setting can be easily adapted to the multi-class setting by using a multi-class model and changing the labelling space and the rewards accordingly. In contrast, extending existing tightly coupled methods in the manners described above would be non-trivial. 

Existing methods typically explore the labelling space in an iterative manner -- in each iteration, they attempt to refine the current labelling in a way that reduces the loss. Because these methods explore the labelling space in a greedy manner, they are very sensitive to initialization. In practice, sophisticated task-specific initialization schemes must be developed in order to achieve good performance with these methods. 

The proposed method takes a different approach. While searching for the best labelling, it maintains an estimate of the region of labelling space that appears promising, which is parameterized by the upper confidence bounds of each arm. This region covers the entire labelling space initially, shrinks over time and converges to the optimal labelling. By maintaining a region rather than a point, the method avoids missing a good labelling that has not yet been explored. By shrinking this region over time, the method avoids wasting time on exploring obviously incorrect labellings. In bandit terms, the method balances exploration with exploitation, enabling a thorough and efficient exploration of the labelling space. Viewed differently from the iterative perspective, the method does not ``commit'' to a labelling in any iteration and only treats the loss obtained in each iteration as one noisy signal, which will be combined with the signals obtained in previous iterations to choose the labelling to try in the next iteration. In practice, the method is able to arrive at a good labelling regardless of the initialization. 

There is a sensible reason that existing methods are tightly coupled -- because strongly supervised models are typically not robust in the presence of significant label noise, extending these to the weakly supervised setting requires learning the model and the labels jointly, so that the model learns to be robust to the modes of noise in the inferred labels. The proposed method is able to decouple label inference and model learning by using the sensitivity of the model to label noise as training signal. In other words, the model's ability to discriminate the quality of the candidate labelling is used as training signal to improve the accuracy of the labelling. More concretely, if the candidate labelling is poor, then the model will generalize poorly to the held-out set, and so the rewards will be low; consequently, the proposed method will avoid generating similar candidate labellings in future iterations. Furthermore, unlike tightly coupled methods that optimize a loss that indirectly depends on the latent labelling, the proposed method directly optimizes the quality of inferred labels. 

\section{Experiments}

\subsection{Binary MIL}

We first compare the proposed algorithm with existing approaches in the standard binary MIL regime. We use the datasets from \cite{andrews2002}, which arise from the drug activity prediction, image classification and text categorization settings and have become the standard benchmarks for evaluating MIL methods. For comparability with the methods introduced in \cite{andrews2002}, we used a vanilla SVM classifier with an RBF kernel to compute the rewards used by the proposed algorithm. 

We choose a reward function that penalizes labellings that cannot be modelled by the classifier or cause violations of the MIL constraints on the held-out set. More formally, let $B(x)$ denote the bag that contains instance $x$, $L_{B} \in \{0,1\}$ denote the label of bag $B$, $\pi(x) \in \{0,1\}$ denote the label assigned to instance $x$, $f(x) \in \{0,1\}$ denote the label of instance $x$ predicted by the strongly supervised classifier, $N(x)$ denote the set of $k$-nearest neighbours of instance $x$ in the held-out set in the space of the classifier output and $\mathbf{1}\left[\cdot\right]$ denote the indicator function. The reward function $R_{x}(\pi(x))$ for assigning the label $\pi(x)$ to the instance $x$ is:

\vspace{-5mm}
\footnotesize
\begin{equation*}
\begin{split}
R_{x}(\pi(x)) & = \mathbf{1}\left[\pi(x)=f(x)\right]\cdot\tilde{R}_{x}(\pi(x))\mbox{, where } \\
\tilde{R}_{x}(\pi(x)) & = \gamma\left(\frac{1}{k}\sum_{x'\in N(x)}Rec(B(x'))\right)+(1-\gamma) \cdot \mathbf{1}\left[\frac{1}{k}\sum_{x'\in N(x)}Rec(B(x'))\geq\alpha\right] \left(\frac{1}{k}\sum_{x'\in N(x)}Prec(x')\right) \\
Rec(B) & = \begin{cases}
\mathbf{1}\left[\exists x\in B\;f(x)=1\right] & L_{B}=1\\
1 & L_{B}=0
\end{cases}
\mbox{ and }
Prec(x)=\begin{cases}
\mathbf{1}\left[L_{B(x)}=1\right] & f(x)=1\\
1 & f(x)=0
\end{cases}
\end{split}
\end{equation*}
\normalsize
Conceptually, $Rec(B)$ is the recall component that penalizes positive bags that don't contain a positive instance, $Prec(x)$ is the precision component that penalizes positive instances that appear in negative bags, $\alpha$ is the minimum average recall level at which the precision reward starts to apply and $\gamma \in (0,1)$ balances the precision and recall components of the reward. 

In our experiments, we used $\alpha = 1$ and $\gamma = 1/7$. In accordance with \cite{andrews2002}, we perform 10-fold cross-validation and report the mean and standard deviation of the bag-level accuracy over ten runs in Table \ref{tab:binary_mil}. As shown, the proposed method combined with a vanilla SVM classifier is able to achieve consistently competitive performance compared to methods designed specifically for the MIL regime. In particular, on some of the text categorization datasets, the proposed method outperforms existing methods by a non-negligible margin, which could possibly indicate that the proposed method is able to avoid getting stuck in the local optima that existing methods are trapped in. 

\begin{table}
\centering
\footnotesize
\begin{tabular}{l m{2cm} m{2cm} m{2cm} m{2cm} }
\toprule 
Dataset & EM-DD \cite{zhang2001dd} & mi-SVM with RBF kernel \cite{andrews2002} & MI-SVM with RBF kernel \cite{andrews2002} & BLISS+SVM w/ RBF kernel \\
\midrule
MUSK1 & $84.8$ & $87.4$ & $77.9$ & $84.2\pm2.5$ \\
MUSK2 & $84.9$ & $83.6$ & $84.3$ & $83.3\pm1.8$ \\
\midrule
Elephant & $78.3$ & $80.0$ & $73.1$ & $83.3\pm1.2$ \\
Fox & $56.1$ & $57.9$ & $58.8$ & $60.0\pm2.2$ \\
Tiger & $72.1$ & $78.9$ & $66.6$ & $79.4\pm1.4$ \\
\midrule
TST1 & $85.8$ & $90.4$ & $93.7$ & $96.0\pm0.2$ \\
TST2 & $84.0$ & $74.3$ & $76.4$ & $74.6\pm1.1$ \\
TST3 & $69.0$ & $69.0$ & $77.4$ & $87.8\pm0.7$ \\
TST4 & $80.5$ & $69.9$ & $77.3$ & $84.5\pm0.5$ \\
TST7 & $75.4$ & $81.3$ & $64.5$ & $80.3\pm1.3$ \\
TST9 & $65.5$ & $55.2$ & $57.0$ & $68.4\pm1.0$ \\
TST10 & $78.5$ & $52.6$ & $69.1$ & $77.8\pm0.8$ \\
\bottomrule
\end{tabular}
\caption{Bag-level accuracy over ten runs on standard binary MIL datasets}
\label{tab:binary_mil}
\end{table}

\subsection{Multi-Class MIL}

We consider a natural extension of the MIL regime to the multi-class setting. Rather than having a single binary label, each bag is now associated with a set of positive labels. It is known that a bag must contain at least one instance of each label in the label set and only instances whose labels are in the label set and negatively-labelled instances. Note that the binary setting is a special case of this multi-class setting where the label sets of positive and negative bags are the singleton set consisting of the unique positive label and the empty set respectively. 

The proposed algorithm can be easily adapted to the multi-class setting by using a multi-class classifier and generalizing the reward function. Specifically, the reward function remains the same as in the binary case except for the following redefinitions:
\footnotesize
\[
Rec(B)=\begin{cases}
\frac{\left|\{l\in L_{B}\left|\exists x\in B\;f(x)=l\right.\}\right|}{\left|L_{B}\right|} & L_{B}\neq\emptyset\\
1 & L_{B}=\emptyset
\end{cases}\mbox{ and }Prec(x)=\begin{cases}
\mathbf{1}\left[f(x)\in L_{B(x)}\right] & f(x)>0\\
1 & f(x)=0
\end{cases}
\]
\normalsize
where $L_{B}$ is now the label set of bag $B$, $f(x) \in \{0,\ldots,M-1\}$ is the prediction of a multi-class classifier, with $0$ denoting the negative class. We also redefine $N(x)$ to be the set of $k$-nearest neighbours of $x$ along the dimension of the classifier output corresponding to the predicted class of $x$. 

Because instances in the negative class typically have multiple modes and are often not linearly separable from the positive classes as a whole, we further extend the algorithm configuration above by introducing multiple negative labels. Since the mode each negative instance belongs to is unknown, we simply include all these negative labels in the set of possible labels for each instance. Because the reward function penalizes labellings that cannot be modelled by the classifier, if a linear classifier is used, only the negative instances that are linearly separable from the positive classes will be assigned the same negative label and so different negative labels tend to capture different modes. As a result, under this configuration, the algorithm is able to learn the different modes of the negative instances in an unsupervised manner. 

We use a modified version of the softmax classifier that does not optimize for discrimination between different negative classes, which will be referred to as the cooperative softmax classifier and has the following objective: 
\footnotesize
\[
\max_{\{w_{j}\}_{j}}\sum_{x\in S}\sum_{l}\mathbf{1}\left[\pi(x)=l\right]\log\sigma_{l}(x,\{w_{j}\}_{j})\mbox{, where }\sigma_{i}(x,\{w_{j}\}_{j})=\frac{\exp(w_{i}^{T}x)}{\exp(w_{i}^{T}x)+\sum_{C\in\mathcal{C}:i\notin C}\max_{j\in C}\left\{ \exp(w_{j}^{T}x)\right\} }
\]
\normalsize
In the formula above, $\mathcal{C}$ is a disjoint and exhaustive grouping of classes such that classes within a set $C \in \mathcal{C}$ do not compete with each other. In our case, $\mathcal{C}$ consists of a set containing all the negative classes and multiple singleton sets, each containing one positive class. 

For our experiments, we constructed a multi-class MIL dataset from the MNIST handwritten digits dataset. Each bag in the constructed dataset contains between five and fifteen randomly chosen instances from the MNIST training set. Digits 0 to 4 are treated as positive classes, with the remaining classes combined into a single negative class. The label set of each bag reflects the presence of instances belonging to positive classes within the bag. The statistics of the constructed dataset are shown in Table \ref{tab:multiclass_mil_data_stats}. 

\begin{table}
\centering
\footnotesize
\begin{tabular}{cccccccc}
\toprule 
Size of Label Set & 0 & 1 & 2 & 3 & 4 & 5 & Total\\
\midrule
Number of Bags & 31 & 322 & 1200 & 2024 & 1795 & 622 & 5994\\
\bottomrule
\end{tabular}
\caption{Statistics of multi-class MIL dataset constructed from MNIST}
\label{tab:multiclass_mil_data_stats}
\end{table}

We ran the proposed algorithm under the configuration described above on this dataset, with the hyperparameters selected based on the bag-level recall and precision on the validation set. As a baseline, we also trained mi-SVM, which only works in the binary setting, in a one-vs-rest manner and compare it to the proposed algorithm. The instance-level accuracy achieved by both methods on the MNIST training and test sets as well as the accuracy of labels inferred by the proposed algorithm are reported in Table \ref{tab:multiclass_mil}. 

\begin{table}
\centering
\footnotesize
\begin{tabular}{lccccc}
\toprule 
 & \multicolumn{2}{c}{mi-SVM w/ linear kernel}  & \multicolumn{3}{c}{BLISS+Cooperative Softmax} \\
 \cmidrule(r){2-6}
 & Train & Test & Label Inference & Train & Test \\
\midrule
Negative Classes (5 - 9) & 29.2 & 31.6 & 50.3 & 46.3 & 47.9\\
Positive Class 0 & 98.6 & 99.1 & 96.8 & 96.9 & 97.9\\
Positive Class 1 & 98.9 & 99.5 & 96.1 & 97.6 & 98.1\\
Positive Class 2 & 89.7 & 91.0 & 89.1 & 88.3 & 89.1\\
Positive Class 3 & 92.4 & 93.6 & 89.6 & 89.7 & 92.9\\
Positive Class 4 & 96.6 & 96.6 & 95.1 & 95.2 & 95.9\\
\midrule
Overall & 62.9 & 64.7 & 72.3 & 70.4 & 72.0\\
\bottomrule
\end{tabular}
\caption{Instance-level accuracy on MNIST training and test sets}
\label{tab:multiclass_mil}
\end{table}

As shown, both methods have difficulty disentangling negative instances from positive instances, which is not surprising since only $0.5\%$ of the bags are negative bags, i.e.: those that contain only negative instances. However, the proposed algorithm produces far fewer false positives compared to mi-SVM while generating only slightly more false negatives, suggesting that it better models the negative instances. Overall, the proposed algorithm improves instance-level accuracy on the test set by $7.3\%$ compared to mi-SVM. 

\subsection{MIL with Domain-Specific Prior}

We apply the proposed algorithm to the task of object detection, the goal of which is to predict the locations of bounding boxes of objects in a category of interest in an image. Because manually annotating the bounding boxes of objects in images is labour-intensive and costly, we would like to leverage the plethora of images with only image-level category labels that are available online to train an object detector. This naturally gives rise to a multi-class MIL problem, where the bags correspond to images and instances correspond to bounding boxes in images that could plausibly contain objects. Positive labels correspond to different foreground object categories and the negative label corresponds to background. 

For comparability with existing MIL-based methods like \cite{song2014learning}, we used the same preprocessing pipeline to extract bounding box proposals from images in the PASCAL VOC 2007 dataset and compute features on the bounding boxes. The resulting training set consists of 5011 bags, each containing on the order of 2000 instances on average. Due to the size of the dataset, we first eliminate the obvious negative instances and split each bag with multiple positive labels into several smaller bags. We do so by running the proposed algorithm with the cooperative softmax classifier and multi-class MIL rewards, with the hyperparameters set to ensure high bag-level recall and reasonable bag-level precision. Then for each original bag, we construct smaller bags consisting of only the instances in the original bag that have the same positive inferred label, thereby reducing the multi-class MIL problem to a binary MIL problem. Next, we run the proposed algorithm again with the cooperative softmax classifier and binary MIL rewards augmented with a domain-specific prior that captures the intuition that if an instance is positive, there are instances in other positive bags that are similar and no instances in negative bags that are similar. 

\begin{figure}[h]
    \centering
    \includegraphics[width=1.0\textwidth]{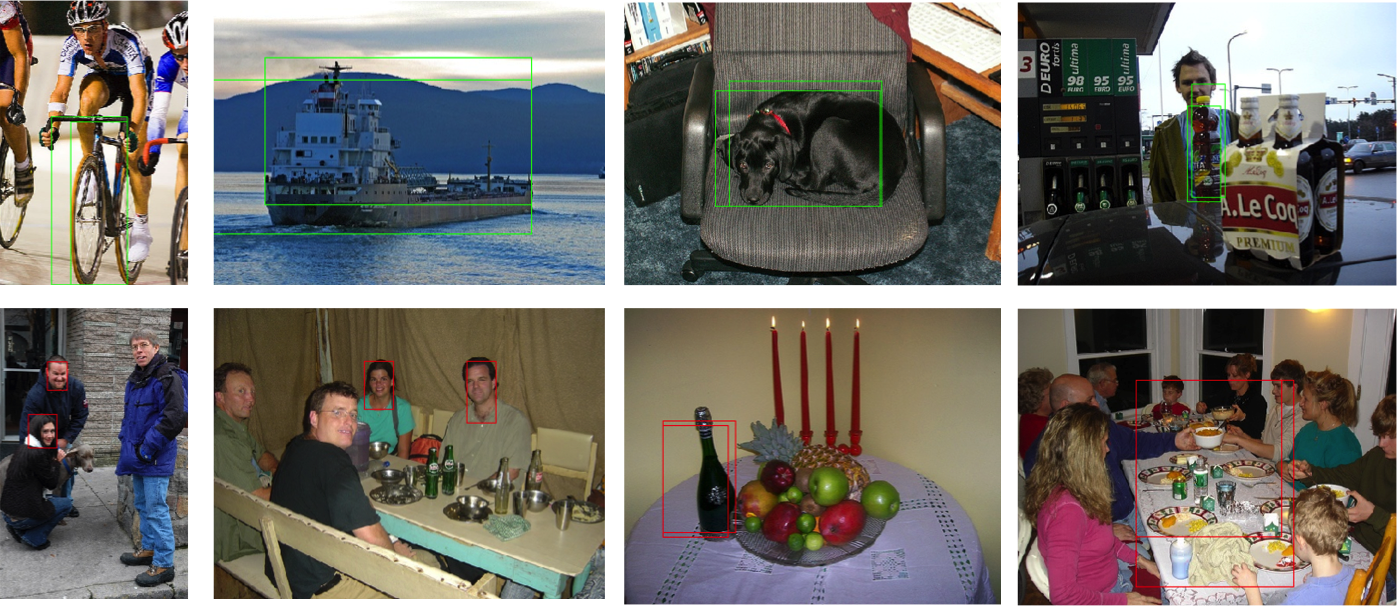}
    \label{fig:inferred_labels}
    \caption{Examples of correct and incorrect instances with positive inferred labels on PASCAL VOC 2007 trainval set. An instance with a positive inferred label is considered correct if the bounding box it is associated with overlaps with the ground truth bounding box by more than 50\%, where overlap is defined as the intersection over union (IoU) between the bounding boxes. }
\end{figure}

Formally, let $\left.N(x)\right|_{B}$ denote the set of $k$-nearest neighbours of $x$ in the bag $B$ in the space of classifier output, $\mathcal{B}_{l}$ be the set of bags with the positive label $l$, $\eta(\cdot)$ be a function that clips and normalizes the value of its parameter to lie within $[0,1]$ and $\tilde{R}_{x}'(\cdot)$ denote the original unaugmented $\tilde{R}_{x}(\cdot)$ defined above. We use the following augmented reward function:
\footnotesize
\[
\tilde{R}_{x}(\pi(x))=\begin{cases}
DistGap(B(x),x)\cdot\tilde{R}_{x}'(\pi(x)) & \pi(x)>0\\
\left(1-DistGap(B(x),x)\right)\cdot\tilde{R}_{x}'(\pi(x)) & \pi(x)\leq0
\end{cases}\mbox{,}
\]
\[
\mbox{where }DistGap(B,x)=\eta\left(\frac{1}{k\left|\mathcal{B}_{L_{B}}^{c}\right|}\sum_{B'\in\mathcal{B}_{L_{B}}^{c}}\sum_{x'\in\left.N(x)\right|_{B'}}\left\Vert x'-x\right\Vert _{2}-\frac{1}{k\left|\mathcal{B}_{L_{B}}\right|}\sum_{B'\in\mathcal{B}_{L_{B}}}\sum_{x'\in\left.N(x)\right|_{B'}}\left\Vert x'-x\right\Vert _{2}\right)
\]
\normalsize 
After labels are inferred, we take the most confident instance with a positive inferred label from each bag, which corresponds to a bounding box, and train an object detector on these inferred bounding boxes using the same procedure and hyperparameters as \cite{song2014learning}. Figure \ref{fig:inferred_labels} shows some instances with correct and incorrect inferred labels. As shown, the proposed algorithm is able to localize objects fairly accurately. In particular, the algorithm localizes faces very well, which is considered incorrect because the ground truth category is person. However, because faces and persons always co-occur, there is in fact no semantic difference between face and person when given only image-level labels. The inferred bounding boxes for bottle and dining table are roughly at the correct locations, but did not capture the full extent of the objects. We report average precision results achieved by the detector trained on the inferred bounding boxes in Table \ref{tab:detection_map}. As shown, the proposed algorithm with random initialization and a simple strongly supervised classifier was able to achieve competitive performance compared to \cite{song2014learning}, which used a sophisticated initialization scheme and a nontrivial extension of MI-SVM. 

\begin{table}
\centering
\footnotesize
\begin{tabular}{lccccccccccc}
\toprule 
 & aero & bike & bird & boat & bottle & bus & car & cat & chair & cow\\
\midrule
MaxCover+SLSVM \cite{song2014learning} & 27.6 & 41.9 & 19.7 & 9.1 & 10.4 & 35.8 & 39.1 & 33.6 & 0.6 & 20.9\\
BLISS+Coop. Softmax & 34.6 & 41.0 & 24.9 & 13.1 & 15.1 & 37.0 & 41.2 & 22.6 & 11.6 & 19.5\\
\midrule
 & table & dog & horse & mbike & person & plant & sheep & sofa & train & tv & mAP\\
 \midrule
MaxCover+SLSVM \cite{song2014learning} & 10.0 & 27.7 & 29.4 & 39.2 & 9.1 & 19.3 & 20.5 & 17.1 & 35.6 & 7.1 & 22.7\\
BLISS+Coop. Softmax & 4.5 & 20.9 & 25.5 & 34.8 & 2.1 & 15.9 & 14.2 & 20.1 & 38.2 & 23.7 & 23.0\\
\bottomrule
\end{tabular}
\caption{Object detection results on PASCAL VOC 2007 test set}
\label{tab:detection_map}
\end{table}

\section{Conclusion}

We presented a general-purpose method for weakly supervised learning that can be applied to any weak supervision regime and enables any strongly supervised classifier to work in the weakly supervised setting. The proposed method decomposes any weakly supervised learning problem into a label inference problem and a strongly supervised learning problem and unifies the disparate weak supervision regimes by representing them simply as user-defined loss functions. We hope this work will encourage exploration of novel weak supervision regimes that are particularly suited for specific domains and enable performance gains achieved under one weakly supervised setting to be easily transferred to other weakly supervised settings. 

\bibliographystyle{plain}
\bibliography{wsl}

\end{document}